\documentclass{article}
\usepackage{spconf,amsmath,graphicx}
\usepackage{times}
\usepackage{epsfig}
\usepackage{graphicx}
\usepackage{amsmath}
\usepackage{amssymb}
\usepackage{units}
\def\etal{\emph{et~al}.}
\newcommand\scalemath[2]{\scalebox{#1}{\mbox{\ensuremath{\displaystyle #2}}}}

\title{Natural Statistics of Network Activations \\ and Implications for Knowledge Distillation}
\name{Michael Rotman and Lior Wolf\sthanks{This project has received funding from the European Research Council (ERC) under the European Unions Horizon 2020 research and innovation program (grant ERC CoG 725974). The contribution of the first author is part of a Ph.D. thesis research conducted at TAU.}}
\address{Tel Aviv University}
\begin{document}
%
\maketitle
\begin{abstract}
In a matter that is analog to the study of natural image statistics, we study the natural statistics of the deep neural network activations at various layers. As we show, these statistics, similar to image statistics, follow a power law. We also show, both analytically and empirically, that with depth the exponent of this power law increases at a linear rate. 

As a direct implication of our discoveries, we present a method for performing Knowledge Distillation (KD). While classical KD methods consider the logits of the teacher network, more recent methods obtain a leap in performance by considering the activation maps. This, however, uses metrics that are suitable for comparing images. We propose to employ two additional loss terms that are based on the spectral properties of the intermediate activation maps.  The proposed method obtains state of the art results on multiple image recognition KD benchmarks. 
\end{abstract}
\begin{keywords}
Knowledge Distillation, Image Statistics
\end{keywords}

\section{Introduction}

The hierarchical structure of Convolutional Neural Networks (CNN) has been lined to their ability to capture the visual world in a way that supports a high degree of invariance to image transformations~\cite{poggio2016visual}. Furthermore, their structure leads to an inductive bias that is especially suitable for reconstructing natural images~\cite{ulyanov2018deep}. It is also known that the activations that are computed in the networks are very effective in the setting of transfer learning, even without further finetuning~\cite{yosinski2014transferable,taigman2014deepface}.

Despite the importance of CNNs and the effectiveness of their intermediate representations, there is little work on the statistical properties of the activation maps. This is in contrast to the significance of the known result in the field of natural image statistics.

One of the hallmarks of the study of natural images is the power-law behavior of natural images. In this work, we show that a similar power-law holds also for the activations obtained at deep layers of CNNs. Moreover, based on spectral analysis considerations, the scaling exponent of the power-law is shown to grow linearly with depth.

Our theoretical results are validated empirically. Additionally, an implication of the study is that when comparing two activation maps, the image norms, such as L1 and L2 are not optimal. We suggest instead to employ two well-known spectral norms. The first is the L1 norm in the spectrum domain. The second is the cross-power term~\cite{rabiner1975theory}.

Knowledge Distillation (KD) is an application in which comparing activation maps is essential. While earlier methods compared normalized logits~\cite{hinton2015distilling}, the more recent methods compare the activation maps after each residual block~\cite{heo2019comprehensive}.

Our experiments demonstrate that the KD method that is based on the spectral norms improves performance when distilling deep ResNets~\cite{he2016deep} to shallower networks, both on CIFAR-100 and on ImageNet.

\section{Related Work}

{\bf Natural Image Statistics\quad} Although natural images can be easily distinguished from one another, they exhibit universality. The power spectrum of an ensemble of images, $P\left( k \right)$, when averaged over rotations, is described as a power law,

\begin{equation}
P\left(\left \vert  k \right \vert \right)  = \left \vert  k \right \vert ^{\alpha} \,,
\label{eq:powerlaw}
\end{equation}
for $\alpha \sim -2$~\cite{field1987relations,tolhurst1992amplitude,ruderman1997origins}. Subsequent works aimed at finding deviations from this power law when parts of the images are scaled or occluded~\cite{millane2003scaling}.

\smallskip
{\bf Knowledge Distillation\quad}
Hinton \etal~\cite{hinton2015distilling} introduced a framework, termed Knowledge Distillation (KD) in which a network is trained with the assistance of a pretrained network with a higher-capacity. By entangling the hidden features of the teacher and student networks, FitNets~\cite{romero2014fitnets} were able to compress deep architectures to thinner ones. Recent works in this field apply different criteria for matching between the hidden feature maps of the two networks, either by applying an additional convolution layer~\cite{heo2019comprehensive,romero2014fitnets} on these maps, or by matching between the correlation~\cite{yim2017gift} or the Jacobian~\cite{srinivas2018knowledge} of the feature maps.

\section{The Power Spectrum of Activations}
\begin{figure}[t]
\begin{center}
   \includegraphics[width=0.8\linewidth]{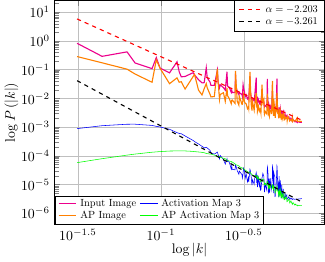}
   \caption{The $\log$-$\log$ plot of the rotational invariant power spectra of the CIFAR100 validation set (magenta) and for the last activation map (blue) as a function of the frequency $\left \vert k \right \vert$. The dashed lines show the slope of the power spectra, $\alpha=-2.203$ for the natural images and $\alpha=-3.261$ for the activation map when fitting to the high-frequency domain of the spectra. The orange and green plots show the power spectra of the average pooled images and activation maps, respectively.}
\label{fig:cifar100powerspectra}
\end{center}
\end{figure}

The rotational invariant power spectrum of the $10000$ images of the CIFAR100~\cite{krizhevsky2009learning} validation set obeys the universal power law in \eqref{eq:powerlaw} as can be seen in Fig.~\ref{fig:cifar100powerspectra}~(magenta).  When transforming back to the spatial domain, the correlation $C\left(r\right)$ between pixels residing at distance $r$ is of the form~\cite{ruderman1997origins}
\begin{equation}
	C\left(r \right) =C_1 + C_2 r^{-\left( 2 + \alpha \right)} \,,
    \label{eq:spatial}
\end{equation}where $C_1$ and $C_2$ are constants. The functional form in Eq.~\eqref{eq:spatial} reveals the scaling properties of natural images. For instance, if an average pooling filter is applied over an image of size $N\times N$ and reduces it to an image of size $\frac{N}{2} \times \frac{N}{2}$, we expect the correlation lengths to decrease by a factor of $2^{2+\alpha}$. However, the power spectrum is invariant (up to the high frequencies, in which data is lost due to the pooling operation) to this scaling as apparent in Fig.~\ref{fig:cifar100powerspectra}~(orange).  

\begin{figure}[t]
\begin{center}
   \includegraphics[width=0.8\linewidth]{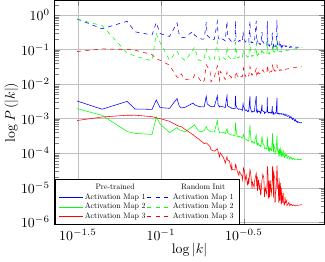}
   \caption{The $\log$-$\log$ plot of the rotational invariant power spectra of the three activation maps of a pre-trained and a randomly initialized (dashed) WideResNet applied on the CIFAR100 validation set as a function of the frequency $\left \vert k \right \vert$. }
\label{fig:actuntrained}
\end{center}
\end{figure}

To further investigate the universal behavior of the activations in a CNN, we examine the power spectrum of the activations of the feature maps residing between the residual blocks of the WideResNet. Fig.~\ref{fig:actuntrained} presents the power spectrum of each activation map for both an untrained network and for a trained network. As can be seen, the activation maps for the untrained map are almost flat, since the parameters of the convolution layers that act on them are sampled from the Gaussian distribution. On the other hand, the activation maps of the trained network exhibit a decay in the high frequency region in the power spectrum as the activations reside in a deeper stage.  This decay is due to the loss of information that occurs between the blocks. A trivial and incorrect explanation would be that the loss of information is a result of the pooling operations. In order to reject this idea, we apply an average pooling operation on the activation map and inspect its power spectrum in Fig.~\ref{fig:cifar100powerspectra}. As can be seen, the high frequency region of the power spectrum also behaves in a universal manner under scaling. Furthermore, The slope in Fig.~\ref{fig:cifar100powerspectra} teaches us about the non-local structure of the activation maps. With an exponent of $\alpha=-3.261$, the correlation length of the activation map increases almost linearly with the distance, pointing to the highly non-local structure of the activation map. Note that activation map $1$ and activation map $2$ in Fig.~\ref{fig:actuntrained} have a higher slope, $\alpha\!=-0.687$ and $\alpha=-1.572$, in their respective power spectra, and therefore exhibit a more localized structure.

\smallskip
\noindent{\bf Analysis\quad}
Consider a $3\times 3$ kernel $W$,
\begin{equation}
W =\scalemath{0.6}{ \left( \begin{array}{ccc} w_{-1,1} & w_{0,1} & w_{1,1} \\  w_{-1,0} & w_{0,0} & w_{1,0} \\ w_{-1,-1} & w_{0,-1} & w_{1,-1}\end{array}\right)} \,.
\end{equation}

Denote by $*$ the convolution operator.  Convolving $W$ with an image $f\left( x \right)$ of size $N\times N$ produces output $g \left(x \right)$ of size $N \times N$, $g\left(x\right) = W * f$.  Using the convolution theorem, the Fourier transform of $g\left(x\right)$,  can be expressed as,
\begin{equation}
    \tilde g \left( k \right)  = \tilde W\left( k \right)  \tilde f \left( k \right) \,,
    \label{eq:convolvefourier}
\end{equation}
where $\tilde W\left( k \right)$ is obtained by zero-padding $W$ to size $N\times N$, and applying the Fourier transform on the zero-padded kernel. The Fourier transform, $\tilde W\left(k\right)$, consists of exactly nine terms,
\begin{equation}
    \tilde W \left( k_x,k_y \right) = \mathcal{N}\sum_{x=-1}^1 \sum_{y=-1}^1 e^{i \left( k_x x + k_y y \right)}w_{x,y} \,,
    \label{eq:fourierW}
\end{equation}
where $k_x$ and $k_y$ are the coordinates in Fourier space and $x$ and $y$ are the coordinates in the spatial domain (the center of the image is located at $x=y=0$). $\mathcal{N}$ is a normalization constant. The polar coordinates is a natural choice for representing the rotation invariant power spectra, $\left(k_r,k_\theta\right) = \left(\sqrt{k_x^2 + k_y^2} \equiv \vert k \vert,  \tan^{-1}\left(\frac{k_y}{k_x}\right)\right)$.
The rotational averaged Fourier transform of the kernel,  $\tilde W \left( \vert k \vert \right)$, only depends on $\vert k \vert$ and consists of three frequency modes, 
\begin{equation}
    \tilde W \left( \vert k \vert \right)= w_{0,0} + e^{i\vert k \vert}W_1 +  e^{i\sqrt{2}\vert k \vert}W_{\sqrt{2}}
\end{equation}
where $W_1 = w_{0,1}+w_{0,-1}+w_{1,0}+w_{-1,0}$ and $W_{\sqrt{2}} = w_{1,1}+w_{1,-1}+w_{-1,1}+w_{-1,-1}$. The three frequencies correspond to the the distances $0$, $1$, and $\sqrt{2}$ from the origin of the elements of $W$. Since the input image $f\left(x\right)$ is isotropic, and as a consequence so is its Fourier transform $\tilde f\left( k\right)$, the power spectrum $P$  is also rotationally invariant, and does not depend on $k_\theta$. Since the power spectrum is averaged over frequencies of the same length, one has to also factor in the Jacobian, of the coordinates transformation which is $\vert k \vert$. This results in a power spectrum of the form $    P\left( \vert k \vert \right) = \vert k \vert \left \vert \tilde W \left( \vert k \vert \right) \right \vert^2 \left \vert \tilde f \left( \vert k \vert \right) \right \vert^2$, Thus contributions to  $\log P\left( \vert k \vert \right)$ are due to the original power spectrum of the input, together with contributions from $\left \vert \tilde W \left( \vert k \vert \right) \right \vert^2$. Since universality only depends on the logarithm of the power spectrum, $P\left( \vert k \vert \right)$, up to multiplicative constants is
\begin{eqnarray}
    P\left( \vert k \vert \right) &\sim&  \vert k \vert \left(w_{0,0}^2 + W_1^2 + W_{\sqrt{2}}^2 \right)\left \vert \tilde f \left( \vert k \vert \right) \right \vert^2 + \nonumber  \\
    && + \vert k \vert  W_1W_{\sqrt{2}} \cos \left( \left(1 - \sqrt{2}\right)\vert k \vert \right)\left \vert \tilde f \left( \vert k \vert \right) \right \vert^2 \nonumber \\
    && + \vert k \vert  w_{0,0}W_{1} \cos \left( \vert k \vert \right)\left \vert \tilde f \left( \vert k \vert \right) \right \vert^2 \nonumber \\
    && + \vert k \vert  w_{0,0}W_{\sqrt{2}} \cos \left( \frac{1}{\sqrt{2}}\vert k \vert \right)\left \vert \tilde f \left( \vert k \vert \right) \right \vert^2 \,. 
    \label{eq:totalpower}
\end{eqnarray}
The first term in Eq.~\eqref{eq:totalpower} comes from the power spectrum of the original image. The other remaining terms are ``interference'' elements that appear due to the frequency content of $\tilde W$. The cosine contribution in these terms is in the range $\left[0,1 \right]$ for the possible values of $\vert k \vert$, and therefore, their contribution to $\log P\left(\vert k \vert  \right)$ is negligible for low frequencies, and becomes dominant for high frequencies. 

The same analysis can be extended to multiple convolutional layers. In this scenario, when multiple layers are applied on an input image $f$, the resulting power spectrum gains a multiplicative contribution from each layer. The logarithm of the power spectrum therefore gains only an additive contribution that is proportional to the number of layers. 

This result shows that the application of multiple convolutional layers only influences the high frequency region of the power spectrum. Furthermore, this analysis explains the empirical behavior of the power spectrum as a function of the activation map depth that is seen in Fig.~\ref{fig:actuntrained}.   

Another issue that surfaces when examining the power spectra structure in Fig.~\ref{fig:actuntrained}, is that unlike the spectra of untrained networks, as the activation map is from a deep layer, its content no longer obeys the Gaussian distribution, and therefore might not be balanced around the mean. As a consequence, the MSE metric may perform poorly since it estimates the mean of the distribution. Combining this with our analysis that each layer obeys a different power law, might indicate that an additional distance metric is required.

\begin{table}[t]
\begin{center}
\footnotesize 	
\begin{tabular}{|l|c|c|c|}
\hline
Setup & Teacher & Student & $\nicefrac{\vert{\Theta_S}\vert}{\vert{\Theta_T}\vert}$\\
\hline\hline
a & WideResNet 28-4&  WideResNet 16-4 &$47.2\%$\\ 
b & WideResNet 28-4 &WideResNet 28-2 &$25.0\%$\\
c & WideResNet 28-4 & WideResNet 16-2 & $11.9\%$\\
d & WideResNet 28-4 & ResNet 56 &$14.7\%$\\
e & PyramidNet-200 &  WideResNet 28-4 & $21.9\%$\\
f & PyramidNet-200 & PyramidNet-100 & $14.6\%$\\
\hline
\end{tabular}
\end{center}
\caption{The teacher-student setups for CIFAR100.}
\label{tab:cifar100networks}
\end{table}

\begin{table*}[t]
\begin{center}
\footnotesize 	
\begin{tabular}{|l|c|c|c|c|c|c|c|c|c|c|}
\hline
Setup &Teacher&Baseline & KD~\cite{hinton2015distilling} & FitNets~\cite{romero2014fitnets}& AT~\cite{zagoruyko2016paying} & Jacobian~\cite{srinivas2018knowledge} & FT~\cite{kim2018paraphrasing} & AB~\cite{heo2019knowledge} & Overhaul\cite{heo2019comprehensive} & Ours \\
\hline\hline
a & 21.09&22.72& 21.69 & 21.85 & 22.07 & 22.18 & 21.72 & 21.36 &20.72&  {\bf 20.37}\\ 
b & 21.09&24.88& 23.43 & 23.94 & 23.80 & 23.70 & 23.41 & 23.19 &22.15 &{\bf 21.45} \\
c & 21.09&27.32& 26.47 & 26.30 & 26.56 & 26.71 & 25.91 & 26.02 & {\bf 24.27} & 24.42 \\
d & 21.09&27.68& 26.76 & 26.35 & 26.66 & 26.60 & 26.20 & 26.04 &25.11 & {\bf 24.87} \\
e & 15.57&21.09& 20.97 & 22.16 & 19.28 & 20.59 & 19.04 & 20.46 &18.03 &  {\bf 17.99} \\
f & 15.57&22.58& 21.68 & 23.79 & 19.93 & 23.49 & 19.53 & 20.89 &19.07 & {\bf 18.67} \\
\hline
\end{tabular}
\end{center}
\caption{Error rates on the CIFAR-100 validation set. Baseline=no distillation. Results for the literature methods are from~\cite{heo2019comprehensive}.}
\label{tab:cifar100results}
\end{table*}
\begin{table}[t]
\begin{center}
\footnotesize 	
\begin{tabular}{|l|c|c|}
\hline 
Method& Err 1 & Err 5 \\
\hline\hline
Teacher & 23.84 & 7.14 \\
Baseline  & 31.13& 11.24 \\
KD~\cite{hinton2015distilling} & 31.42& 11.02 \\
AT~\cite{zagoruyko2016paying} & 30.44& 10.67\\
FT~\cite{kim2018paraphrasing} & 30.12 & 10.50\\
AB~\cite{heo2019knowledge} & 31.11 & 11.29 \\
Overhaul~\cite{heo2019comprehensive} &28.75 & 9.66 \\ 
Ours &  {\bf 27.49} & {\bf 8.99} \\
\hline
\end{tabular}
\end{center}
\caption{Classification error rate on the ILSVRC 2012 validation set. Baseline represents no distillation. Results for other methods are from \cite{heo2019comprehensive}. $\mathcal{T}$ is a ResNet 50 and $\mathcal{S}$ is MobileNet.}
\label{tab:imagenetresults}
\end{table}
\section{Knowledge Distillation}
Let $\left\{\left(x_t,y_t\right) \right\}_{t=1}^n$ be a set of tuples, each contains an example, $x_t\in\mathbb{R}^d$, with its corresponding label, $y_t\in\left\{1,\dots,k\right\}$. Given a pre-trained teacher network, $\mathcal{T}$, with parameters $\Theta_\mathcal{T}$, the objective is to train a student network, $\mathcal{S}$, with parameters $\Theta_\mathcal{S}$, such that $\vert \Theta_\mathcal{S} \vert < \vert \Theta_\mathcal{T} \vert$.

In order to leverage the information encapsulated inside $\mathcal{T}$, a feature-wise term is added to the loss function, so the features of $\mathcal{S}$ are entangled with the features of $\mathcal{T}$. Denote by $F^{\mathcal{T}}_i$ and $F^{\mathcal{S}}_j$ the $i$th feature map in $\mathcal{T}$ and the $j$th feature map in $\mathcal{S}$ (outputs of the $i$th and $j$th convolution layer in $\mathcal{T}$ and $\mathcal{S}$). Assume that $F^{\mathcal{T}}_i$ and $F^{\mathcal{S}}_j$ represent the same embedding of the input $x_t$. Since these feature maps may not share the same dimensionality, learnable transformations $T_{\mathcal{T}}$ and $T_{\mathcal{S}}$ are applied on the feature maps to produce reduced feature maps, each with $M$ channels, denoted next by the index $m$. Once the reduced feature maps, $R^T_{i,m} \equiv T_{\mathcal{T},m}\left(F^{\mathcal{T}}_i \right)_m$ and $R^S_{j,m} \equiv T_{\mathcal{S},m}\left(F^{\mathcal{S}}_j \right)_m$ are embedded in the same space, a similarity metric can be utilized.

Entangling between the reduced feature maps is achieved by introducing a pixel-wise distillation loss term~\cite{heo2019comprehensive}, $\mathcal{L}_{\text{overhaul}}$. This term drives the convergence of the positive pre-ReLU entries in the feature maps of $\mathcal{S}$ towards the feature maps of $\mathcal{T}$. This term, however, only acts on specific entries in the spatial domain of the feature maps, and is unable to capture non-local aspects of the feature maps. To remedy this, two Fourier terms are added to the loss function since non-local properties are naturally captured in Fourier space as in Eq.~\eqref{eq:fourierW}. The first term, $L1$ loss over the Fourier transform of the reduced feature maps, $\mathcal{L}_{L1} = \left\vert \tilde R^T_{i,m}\! -\! \tilde R^S_{j,m} \right\vert $, increases the robustness of the student feature maps in Fourier space. The second term,
\begin{equation}
    \mathcal{L}_{CPS} = \frac{1}{M}\sum_m\left\langle 1-\frac{P^{\mathcal{T}\mathcal{S}}_{ij,m}\left(\vert k \vert\right)}{P^{\mathcal{T}\mathcal{T}}_{ij,m}\left(\vert k \vert\right)P^{\mathcal{S}\mathcal{S}}_{ij,m}\left(\vert k \vert\right)} \right\rangle \,,
\end{equation}
is a cross-power spectrum loss function, where $\langle \cdot \rangle$ denotes expectation over equal $\vert k \vert$ lengths, and $P^{\mathcal{X}\mathcal{Y}}\left(\vert k \vert\right)_{ij,m}$ is,
\begin{equation}
    P^{\mathcal{X}\mathcal{Y}}_{ij,m}\left(\vert k \vert\right) =  \left.{\tilde R^{\mathcal{X}}_{i,m}}\right.^*\left(\vert k \vert \right)\tilde R^{\mathcal{Y}}_{j,m} \left(\vert k \vert \right) \,.
\end{equation}
This term matches between the rotational invariant power spectra of the teacher and the student networks, enforcing the activations of these two networks to share the same non-local structure. The total loss function is
\begin{equation}
    \mathcal{L} = \mathcal{L}_{CE} + \alpha\mathcal{L}_{\text{overhaul}} +  \beta\mathcal{L}_{L1}+ \gamma\mathcal{L}_{CPS} \,,
\end{equation}
where $\mathcal{L}_{CE} = -\sum_{c=1}^k \delta_{c,y_t}\log p_c\left(x_t\right) $ is the cross entropy loss function, and $p_c$ is the probability that $x_t$ belongs to class $c$, assigned by network $\mathcal{S}$.

\section{Experiments}
We show the benefits of combining the spectral information during the process of KD for the task of image recognition. In all experiments, we applied the KD loss terms on the feature maps located after the bottlenecks of $\mathcal{T}$ and $\mathcal{S}$. 

\smallskip
{\bf CIFAR-100\quad}
The CIFAR-100~\cite{krizhevsky2009learning} dataset consists of $60,000$ $32\times32$ color images divided into $100$ classes. There are $50,000$ training examples and $10,000$ for validation. In order to show the importance of spectral matching, our method is validated over several teacher-student setups. The WideResNet~\cite{zagoruyko2016wide} with $28$ hidden layers and $\times4$ channel ratio and PyramidNet-200~\cite{han2017deep} with $240$  hidden layers were used as the teacher networks. For the student networks, smaller versions of the WideResNet, ResNet-$56$, and a shallower version of the PyramidNet with $84$ hidden layers were selected, see Tab.~\ref{tab:cifar100networks}. We used the same setup as Heo \etal~\cite{heo2019comprehensive}. All networks were trained for 200 epochs using SGD with a learning rate of $0.1$ and a momentum of $0.9$, $L2$ regularization of $1e^{-4}$ on the network's parameters, $\alpha=\beta=1e^{-4}$ and $\gamma = 0.01$. The learning rate was multiplied by $0.1$ after $100$ epochs and again after $150$ epochs. For setups (a)-(d) a batch size of $128$ was used whereas for setups (e) and (f) a batch size of $64$ was used for memory considerations. 

Our experimental results appear in \ref{tab:cifar100results}. As can be seen, our method outperforms in five out of the six setups. Under the experimental setup (a), both the Overhaul~\cite{heo2019comprehensive} method and our method outperform the teacher network.

\smallskip
{\bf ImageNet\quad}
The ILSVRC 2012~\cite{russakovsky2015imagenet} dataset contains 1.2M training images and 50,000 validation images. These images are cropped to the size of $224 \times 224$ for both training and evaluation. The teacher and student networks for this task are the ResNet 50 and MobileNet~\cite{howard2017mobilenets}. We used the same setup as Heo \etal~\cite{heo2019comprehensive}, an SGD optimizer with a learning rate of $0.1$ and a momentum of $0.9$, $L2$ regularization of $0.0001$ on the network's parameters, $\alpha=\beta=0.00001$ and $\gamma = 0.001$. The student network was trained with a batch size of $256$ for $100$ epochs. The learning rate was reduced by a factor of $0.1$ every $30$ epochs.

A comparison of our approach to recent methods is shown in Tab.~\ref{tab:imagenetresults}. As can be seen, our approach achieves a substantially lower error rate both in the top-1 and top-5 error rates.

\section{Conclusions}

In this work we have explored the power-law property of the activations of deep neural networks. We show that the correlation lengths grow linearly with depth, whereas the activations become more and more concentrated in Fourier space. This behavior indicates an increasing amount of mutual influences between distant image locations,  which matches the shift that occurs with depth from local processing to higher-level semantic information. 

As an immediate application of our study, we prescribe how to utilize the information in Fourier space as a distance metric for activations of deep layers.  When this metric is used for learning, such as in the field of KD,  it leads to an improvement over the state of the art method.

\newpage
\bibliographystyle{IEEEbib}
\bibliography{egbib}

\end{document}